\documentclass[runningheads]{llncs}
\usepackage[T1]{fontenc}
\usepackage{graphicx}
\usepackage[figuresright]{rotating}
\newcommand{\eqnref}[1]{Eq.~(\ref{#1})}

\newcommand{\tableref}[1]{Table~\ref{#1}} 

\newcommand{\figref}[1]{Fig.~\ref{#1}} 
\newcommand{\equref}[1]{Equation~(\ref{#1})}
\usepackage{amsmath,amssymb}
\usepackage{tikz}
\usetikzlibrary{positioning, calc, shapes.geometric, shapes, shapes.multipart, arrows.meta, arrows, decorations.markings, external, trees}
\usepackage{gensymb}
\usetikzlibrary{bayesnet}
\usepackage{booktabs}
\usepackage{hyperref}
\usepackage{subfigure}
\usepackage[misc]{ifsym}
\usepackage{multirow}
\graphicspath{{image/}}
\begin{document}
\title{Route to  Time and  Time to Route: \\ Travel Time Estimation from Sparse Trajectories}
\titlerunning{Route to  Time and  Time to Route}
%
%
\author{Zhiwen Zhang*\inst{2}  \and
Hongjun Wang*\inst{1} \and
Zipei Fan \Letter \inst{1,2}\and \\
Jiyuan Chen\inst{1}\and
Xuan Song \Letter \inst{1} \and
Ryosuke	Shibasaki \inst{2} }
%
%

\institute{  Southern University of Science and Technology, Shenzhen, China  \and
The University of Tokyo, Tokyo, Japan \\
* Equal contribution \\
fanzipei@iis.u-tokyo.ac.jp, songx@sustech.edu.cn }
\maketitle              
\begin{abstract}

Due to the rapid development of Internet of Things (IoT) technologies, many online web apps (e.g., Google Map and Uber) estimate the travel time of trajectory data collected by mobile devices. However, in reality, complex factors, such as network communication and energy constraints, make multiple trajectories collected at a low sampling rate. In this case, this paper aims to resolve the problem of travel time estimation (TTE) and route recovery in sparse scenarios, which often leads to the uncertain label of travel time and route between continuously sampled GPS points. We formulate this problem as an inexact supervision problem in which the training data has coarsely grained labels and jointly solve the tasks of TTE and route recovery. And we argue that both two tasks are complementary to each other in the model-learning procedure and hold such a relation: more precise travel time can lead to better inference for routes (\textit{Time} $\rightarrow $ \textit{Route}), in turn, resulting in a more accurate time estimation (\textit{Route} $\rightarrow $ \textit{Time}). Based on this assumption, we propose an EM algorithm 
to alternatively estimate the travel time of inferred route through weak supervision in $E$ step and retrieve the route based on estimated travel time in $M$ step for sparse trajectories. We conducted experiments on three real-world trajectory datasets and demonstrated the effectiveness of the proposed method.


\end{abstract}
 \section{INTRODUCTION}

With advances in the area of the Internet of Things (IoT), GPS modules have been widely used throughout various kinds of mobile devices. These devices collected massive trajectory data and empowered many applications in the intelligent transportation system. Among these applications, travel time estimation (TTE) is an essential task for route planning, taxi dispatching, and ride-sharing. 
Subsequently, a large part of relevant approaches ranging
from machine learning technologies, such as Bayesian inference \cite{mil2018modified}, to deep learning models \cite{wang2018will} have been proposed to solve this task. 
However, due to the power and communication limitations of the mobile devices, the sampling rate of the trajectories is always low, which leads to a decrease in the accuracy for both travel time and route.  Existing efforts need to label the exact travel time and route between two consecutively sampled GPS points, which is used to train the estimation model. We argue that this hypothesis sounds reasonable relying upon the scene of high-sampling-rate. In practice, we have to face a large part of trajectory data with low sampling rates \cite{lou2009map}.


\begin{figure}[t]
	\centering
	\includegraphics[width=1\linewidth]{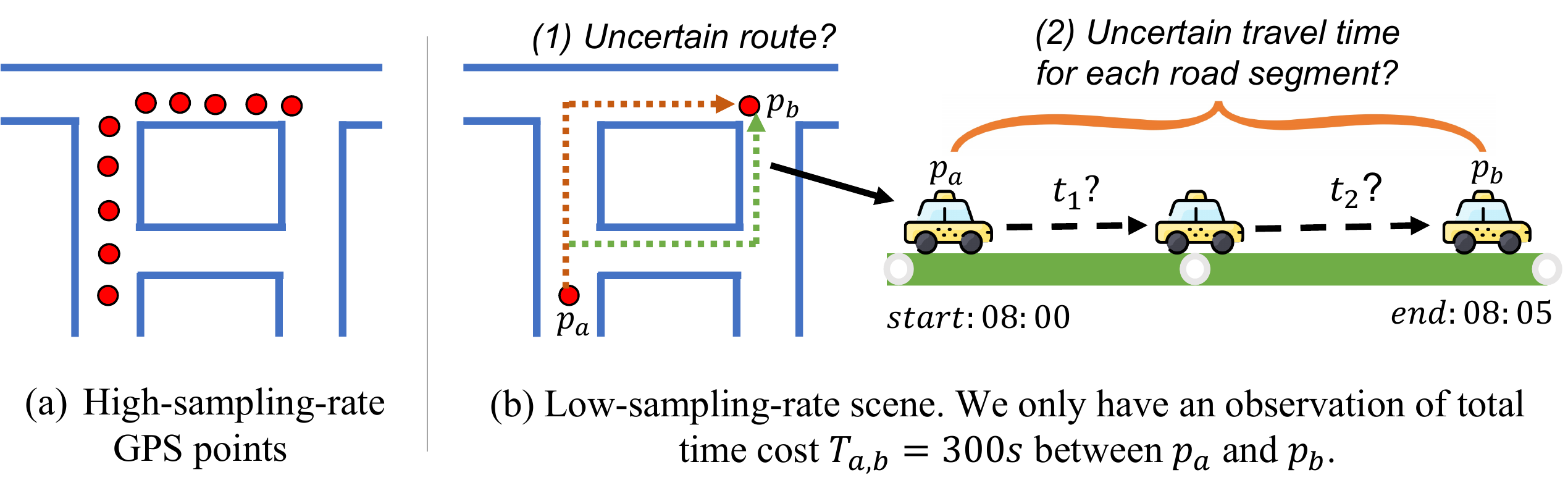}
	\caption{Comparison between dense and sparse TTE scenarios.  } 
	\label{fig:route_choice}
\end{figure}
\textbf{Motivating Scenario.} \figref{fig:route_choice} gives a comparison between two TTE scenarios: dense and sparse. The exact label of travel time in each road segment and the route can be easily obtained from the dense trajectories (\figref{fig:route_choice}.a). However, we can not obtain the precise route and travel time label from the sparse trajectories (perhaps 5-8 minutes for each point). \figref{fig:route_choice}.b illustrates a case in low-sampling rate. We are hard to infer the route when given two sampled GPS points $p_a$ and $p_b$, since there are multiple choices for possible route.
Meanwhile, we were also challenged to acquire the exact travel time in each road segment, even though we have the ground truth route marked with a green dotted line, due to the large gap of observation $T_{a,b}=5$ minute. With missing supervision labels, traditional supervised learning is clumsy  in giving a fine-grained prediction. This motivates us to model TTE and route recovery from sparse trajectories as a weakly supervised learning problem, more specifically a coarse labeling problem \cite{zhang2020learning}.



Unlike conventional supervised learning, where each sample is assigned with a label, coarse labeling annotates the label on a bag of samples. The authors in \cite{zhang2020learning} summarize the task of learning from 1) the  mean/sum: the arithmetic mean or the sum of $X$; 2) the difference/rank: the difference $x_i-x_j$ or the relative order $x_i>x_j$; and 3) the min/max: the smallest/largest value in $X$. In the task of estimating travel time, the problem can be considered as learning from the mean/sum of $X$, since the path travel time can be equivalent to the summation of each road pass time within the path, while the exact travel time of each road segment is unknown.

As we know, travel time and route are highly correlated. In addition, the exact routes can result in a better inference of travel time (\textit{Route} $\rightarrow $ \textit{Time}), in turn leading to a more precise route recovery (\textit{Time} $\rightarrow $ \textit{Route}). In this paper, the Expectation-Maximization (EM) algorithm \cite{dempster1977maximum} has been applied to 
alternatively estimate the travel time and route between any two consecutive GPS points. Technically, the \textbf{E step} intends to estimate the travel time of inferred route through weakly supervised learning (WSL), and the \textbf{M step} schemes to recover the route by heuristically searching for estimated travel time. Furthermore, to model the time-variant representation of road network, we generate the travel time distributions using the proposed spatio-temporal model. The Lognormal distribution is employed in this paper thanks to the excellent nature of additivity \cite{dufresne2008sums} and better performance in fitting real travel time.


The main contributions of this paper can be summarized as follows.
\begin{itemize}
	\item[$\bullet$] For the first time, we integrate weakly supervised learning into the problem of TTE, which aims to infer the travel time of each road segment in
	 a \textit{bag} from a large gap of consecutive GPS points.
	\item[$\bullet$] The EM algorithm has been designed to alternatively infer the travel time distribution of each road segment and route between two consistent GPS samples (\textit{Route} $\rightarrow $ \textit{Time} and \textit{Time} $\rightarrow $ \textit{Route}). In addition, we propose a spatio-temporal embedding architecture to forecast the future traffic state that integrates the spatial relational road network and temporal correlations, such as weather conditions and time-of-day.
	\item[$\bullet$] We conduct extensive experiments on three real-world large-scale trajectory data sets, which significantly outperform the state-of-the-art baselines for both two tasks - TTE and route recovery. 
\end{itemize}
\section{ RELATED WORK}

\noindent\textbf{Weakly Supervised Learning.}
Weakly supervised learning focuses on dealing with three kinds of problems \cite{zhou2018brief}: 1) incomplete supervision: only part of the training data is labeled 2) inexact supervision: training data has only coarsely grained labels 3) inaccurate supervision: given labels are not always accurate. Multiple instance learning (MIL), which deals with observed data arranged in sets \cite{frenay2013classification} is a branch of weakly supervised learning belonging to the category of inexact supervision. MIL has been widely applied in many fields, such as image and video classification \cite{chen2006miles}, as well as document 
and sound classification \cite{zhou2009multi}. This paper expands the concept of MIL to the application of travel time estimation in a highly sparse scenario.

\noindent\textbf{Travel Time Estimation.}
The loop detectors are firstly used in calculating the travel time by recording the individual road travel speeds and dividing it by the travel distance \cite{li2017diffusion}. However, since traffic lights and left/right turns are omitted, the estimation errors are inaccurate. Therefore, road segment-based methods have been proposed, which can be approximately divided into two types: 1) nearest neighbor search \cite{tiesyte2008similarity}, which sets the prediction by averaging the historical trajectory travel time; and 2) trajectory regression methods \cite{ide2011trajectory}, which predict the travel time of road segment by road features. However, those approaches are based on the assumption that the trajectory's travel time is precise. Moreover, multiple trajectories with low sampling rates exist due to the network communication problem. Although some works try to conduct sparse travel time estimation\cite{jabari2020sparse,wang2014travel}, the uncertain route is also ignored. This paper aims to simultaneously resolve the problem of estimating vehicle travel time and route recovery in a highly sparse scenario.

\noindent\textbf{Route Recovery.} 
The route recovery problem in the low sample rate scenario is vital to reduce the uncertainty of the trajectory, and the TTE problem \cite{ren2021mtrajrec}. As we mentioned, the problems of TTE and route recovery play a role together, and this idea has been considered in previous work. For example, \cite{wu2016probabilistic} designs a regression TTE model and applies the exact route search to obtain the potential route based on the learned travel time.  \cite{shao2020estimation} proposes STGAN to generate a travel time distribution in each road segment throughout the road network based on data from traffic surveillance cameras and update the possible route by posterior estimation in every iteration. Unlike the existing approach, the superior advantage of WSL-TTE is to model the sparse observation problem as weakly supervised learning, which is skilled  in coarse labeling problems, and adopt it into the EM framework. 


%
\section{METHODOLOGY}

This section first gives the problem formulation of travel time estimation based on weakly supervised learning and then introduces our proposed weakly supervised learning travel time estimation (WSL-TTE) system.

\begin{figure}[h]
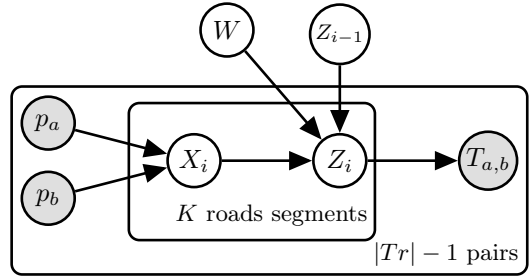

	\centering
	\vspace{-10pt}
	\tikz[line width=1pt]{
		\node[latent] (x) {$X_{i}$};%
		\node[obs, left = of x, xshift=-0.2cm, yshift=0.5cm] (pa) {$p_a$};%

		\node[obs, left = of x, xshift=-0.2cm, yshift=-0.5cm] (pb) {$p_b$};%
		\node[latent, right = of x, xshift=0.2cm] (z) {$Z_{i}$};%
		\node[obs, right = of z, xshift=0.2cm,yshift=-0.0cm] (t) {$T_{a,b}$};%
		\node[latent,  xshift=-1.5cm, above = 10mm of z] (w) {$W$};%
		\node[latent,  xshift=0cm, above = 9mm of z,xshift=0cm, font=\fontsize{8}{6}\selectfont] (zz) {$Z_{i-1}$};%
		\edge {w,x,zz} {z}
		\edge {z} {t}
		\edge {pa,pb} {x}
		
		\plate [inner sep=2.8mm, xshift=-0.2cm, yshift=0.1cm] {plate1} {(x)(z)} {$K$ roads segments}; %
		\plate [inner sep=1.0mm, xshift=-0.0cm] {plate1} {(x)(z)(t)(pa)(pb)} {$|Tr|-1$ pairs}; %
	}
	\caption{ The graphical model of the data generating process. The grey and white nodes represent the observation and hidden variable, respectively. 	}\label{fig:data_gens}
    \vspace{-10pt}
\end{figure}

\subsection{Notation}
Let $X$ be the features of the road (e.g., road types, road lanes) and \textbf{$Z$ be the unobserved true travel-time distribution of $K$ road segments that we want to predict.} The goal is to learn a discriminative model $f$
that predicts the true target $Z$ from the feature vector $X$, so as to maximize the conditional probability $P(Z\mid X)$. Here, the travel time distribution $Z$ highly depends on the real-time traffic condition. Intuitively, we discretize one day into $\mathcal{I}$ time steps (i.e., a specific time window $\Delta t$=30 minutes). 
So given a trajectory $Tr$ at time step $t_s$, which can be denoted as sequence of sampled GPS point: $p_{1} \rightarrow p_{2} \rightarrow \cdots \rightarrow p_{|T r|}$,    $\widetilde{r}$  is the most likely  route (\textit{bag}) between two continuous GPS points $p_a$ and $p_b$ 
during $Tr$, where $|\widetilde{r}|=K$ denotes the number of total travelled road segments in $\widetilde{r}$. Especially,
$X_{1:K}$\footnote{The subscript for example $X_{1:K}$ denotes an abbreviation for the set $\{X_1,X_2,\cdots,X_K\}$} stands for the \textit{bag} of the travelled road segments' features corresponding to $\widetilde{r}$.   $W$ is the collection of all weight parameters in the neural network $f$ to learn $Z$. $T_{a,b}$ is the actual observation of total time cost between $p_a$ and $p_b$.

\figref{fig:data_gens} illustrates the data generation process with a graphical representation. Given $p_a$ and $p_b$, we first infer the 
most likely  route $\widetilde{r}$ from the road network with $P(\widetilde{r}\mid p_a,p_b,Z)$, which is equal to estimate $P(X_{1:K}\mid p_a,p_b,Z_{1:K})$.  Subsequently, we generate the conditional probability  $P\left(Z_i \mid \theta \right)$ for each travel time  distribution $Z_i$.  Specifically, we here assume that each latent variable $Z_i$ belongs to the Lognormal distribution $Z_i \sim \frac{1}{z \sigma \sqrt{2 \pi}} \exp \left(-\frac{(\ln z -\mu)^{2}}{2 \sigma^{2}}\right)$ \cite{pu2011analytic}. We let the conditional probability $P\left(Z_i \mid \theta \right)=P\left(Z_i \mid \left(\mu,\sigma\right)=f\left(X;W\right)\right)$, where $W$ and $X$ are the cofactors that generate the parameterized $\mu$ and $\sigma$ of $Z_i$ by the deterministic function $f$.  Consequently, total travel time 
$T_{a,b}$  can be observed by an aggregate function $Q$ with $T_{a,b}=Q(Z_{1:K})=\sum_{i=1}^{K}Z_i$.

\subsection{Assumptions}
 Here, we summarize the two basic assumptions used in our paper. \\
\noindent\textbf{Assumption 1} \textit{(Aggregate observation assumption)} $P(T_{a,b} \mid X_{1:K}, Z_{1:K}) = P(T_{a,b} \mid Z_{1:K})$

\textit{We assume that  observation $T_{a,b}$ is conditionally independent on $X_{1:K}$ when given $Z_{1:K}$. This assumption is informed by existing studies \cite{carbonneau2018multiple} and conforms to the TTE problem, since given $Z_{1:K}$,  observation $T_{a,b}$ can be determined by  aggregate function $Q$.}

\noindent\textbf{Assumption 2} \textit{(Markov chain assumption)} $P(Z_{1: K} \mid X_{1: K})  =P(Z_{1}\mid X_{1})\prod_{i=2}^{K}    \allowbreak  P(Z_{i+1}   \mid X_{i+1}; Z_{i})$

\textit{We assume that $Z_{i+1}$ are mutually independent except for $Z_{i}$. This is under the assumption of a Markov chain and is based on extensive applications in trajectory data mining \cite{brakatsoulas2005map}. Furthermore, since $T_{a,b}$ can be determined by the function $Q$, the conditional probability can be defined as $P(T_{a,b}\mid Z_{1: K})=\delta_{Q\left(Z_{1: K}\right)}(T_{a,b})$, where $\delta(\cdot)$ represents the \textit{Dirac delta function}.}
\subsection{Problem Formulation}
In summary, the objective function in this paper can be written as follows.
\begin{small}
 	\begin{align}\label{eqn:obj}
 		\log P(T_{a,b} \mid p_a,  p_b,  W)=\log \left(\sum_{Z} P(T_{a,b} \mid p_a,  p_b,  Z,  W) P(Z \mid p_a,  p_b,  W)\right),
 	\end{align}
\end{small}
 which is the maximum of a posterior estimation by taking the $Z$ as latent variables with the observation of sparse travel time $T_{a,b}$. Therefore, we divide the training process into expectation ($E$ step) and maximization ($M$ step) according to the above assumptions.

\begin{align}
 \text{\textbf{E step:}} \quad &\mathbb{E}_{Z}\left[\log P(T_{a,b}, Z \mid X_{1:K}, W)\mid X_{1:K}^{(i)}; W^{(i)} \right]
\nonumber\\
=&\int_{Z^{K}} \log P(T_{a,b}\mid Z_{1:K}; X_{1:K}) P(Z_{1:K}\mid X_{1:K}) d_{Z_{1:K}} \nonumber\\
=&\int_{Z^{K}} \delta_{Q\left(z_{1: K}\right)}(T_{a,b}) 
P(Z_{1} \mid X_{1})\prod_{i=2}^{K}   \allowbreak P(Z_{i+1}   \mid X_{i+1}; Z_{i})d_{Z_{1:K}} \nonumber \quad \\
\approx&\underset{\substack{Z_{i} \sim p\left(Z_{i} \mid X_{i}\right) \\ i=1, \ldots, K}}{\mathbb{E}}\left[\delta_{Q\left(Z_{1: K}\right)} (T_{a,b}) \right]\label{eqn:weakly} \\
\text{} \nonumber \\
\text{\textbf{M step:}} \quad \widetilde{r}& =\arg \max_{r} \log P(T_{a,b} \mid \Omega_{a,b};\ p_a;\ p_b;\ W)\label{eqn:search}
\end{align}
\textbf{E step} aims to estimate the travel time of the most likely  route $\widetilde{r}$ by learned travel time distribution $Z$ through weakly supervised learning, and \textbf{M step} heuristically searches $\widetilde{r}$ from the candidate set $\Omega_{a,b}=\{r_1,r_2,\cdots,r_m\}$ to reduce the computational cost. According to the above EM procedure, we obtain the estimated route $\widetilde{r}$ for every pair of $p_a$ and $p_b$, as well as travel time distribution $Z$. Thus, the final travel time of trajectory $Tr$ can be obtained by summing all the estimation components $\Theta=\{ \widetilde{T}_1,\widetilde{T}_2, \cdots, \widetilde{T}_{|Tr|-1} \}$ between every continuous GPS sample  with $\widetilde{T}=\sum_{i=1}^{|Tr|-1} \widetilde{T}_i$, where $\widetilde{T}_i$, produced by $f$, denotes the forecast travel time of $p_{i}$ and $p_{i-1}$. 



 \begin{figure*}[h]
	\centering
	\includegraphics[width=1\linewidth]{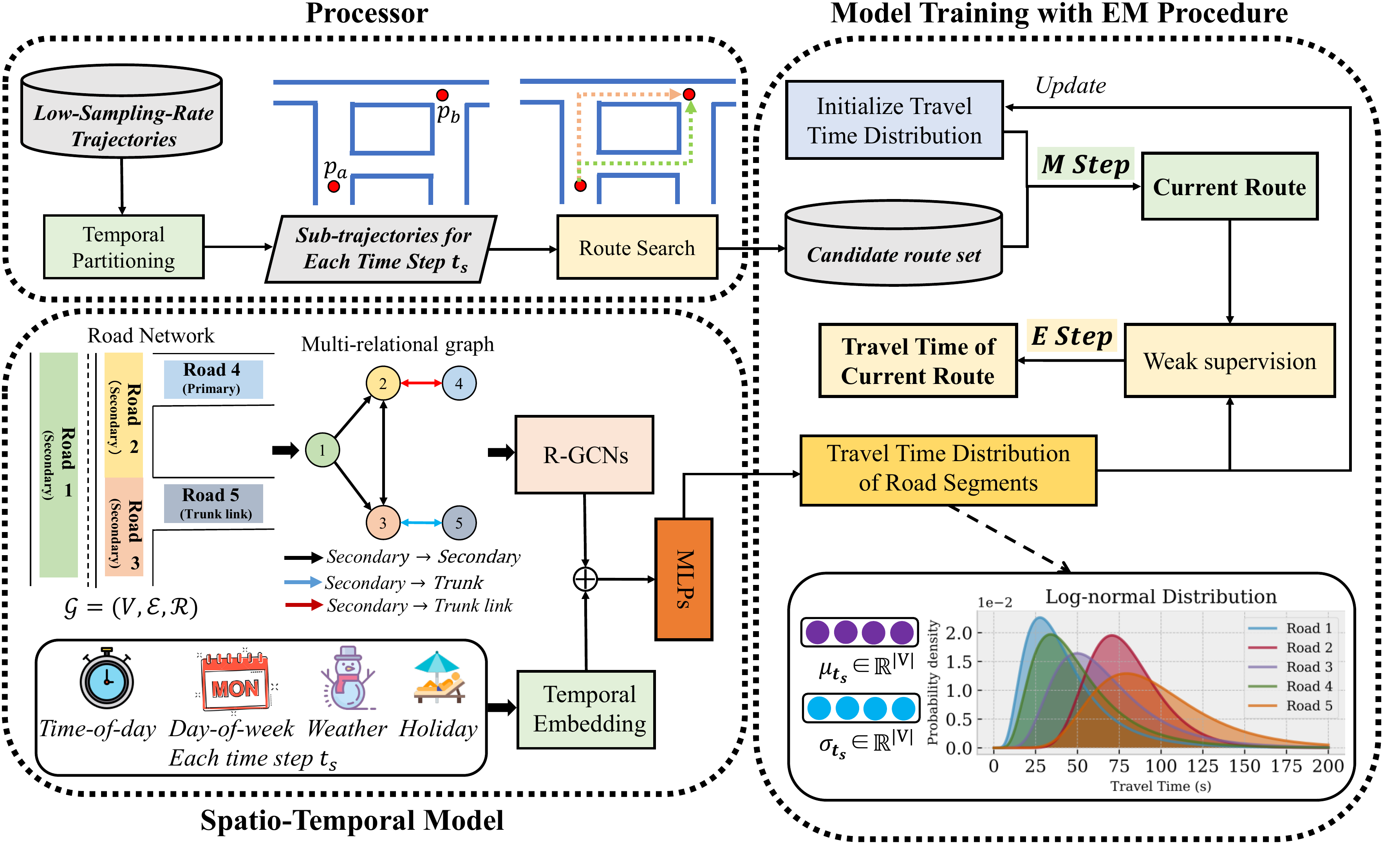}
	\caption{The system architecture of our proposed WSL-TTE with EM procedure. }
	\label{fig:framework}
	\vspace{-10pt}
\end{figure*}

\subsection{System Overview}
 \figref{fig:framework} shows our proposed WSL-TTE system with the EM algorithm, which consists of three main components - processor, spatio-temporal model and model training with EM procedure.
 
 $(1)$ \textbf{Processor} temporally partitions the low-sampling-rate historical trajectories from the datasets into each time step $t_s$.

 
 $(2)$ \textbf{Spatio-Temporal Model} $f$ estimates the travel time distribution $Z$ of the road segments. We firstly transform a road network into a multi-relational graph $\mathcal{G}$, and  encode each time step $t_s$ into a vector. Then we fuse the aforementioned spatial representation and temporal encoding as the time-variant vertex representations. In the final, $\mu_{t_s}$ and $\sigma_{t_s}$ of travel time distribution $Z$ are parameterized by two MLPs (Multi-Layer Perception). 

(3) \textbf{Model Training with EM Procedure} is used to optimize the learned $Z$ in  $E$-step using \eqnref{equ:travel_loss} and  infer route $\widetilde{r}$  by \eqnref{equ:update_route}. 
Travel time distribution $Z$  is initially default  as  $\mu=\frac{L_i}{\mathbf{S}_i}$ and $\sigma=1$, where $L_i$ and $\mathbf{S}_i$ are the length and speed limits of $i^{th}$ road segment, respectively. 
The EM algorithm will be finished when the estimated variables $Z$ converge.


\begin{figure}[!b]
	\centering
	\includegraphics[width=0.8\linewidth]{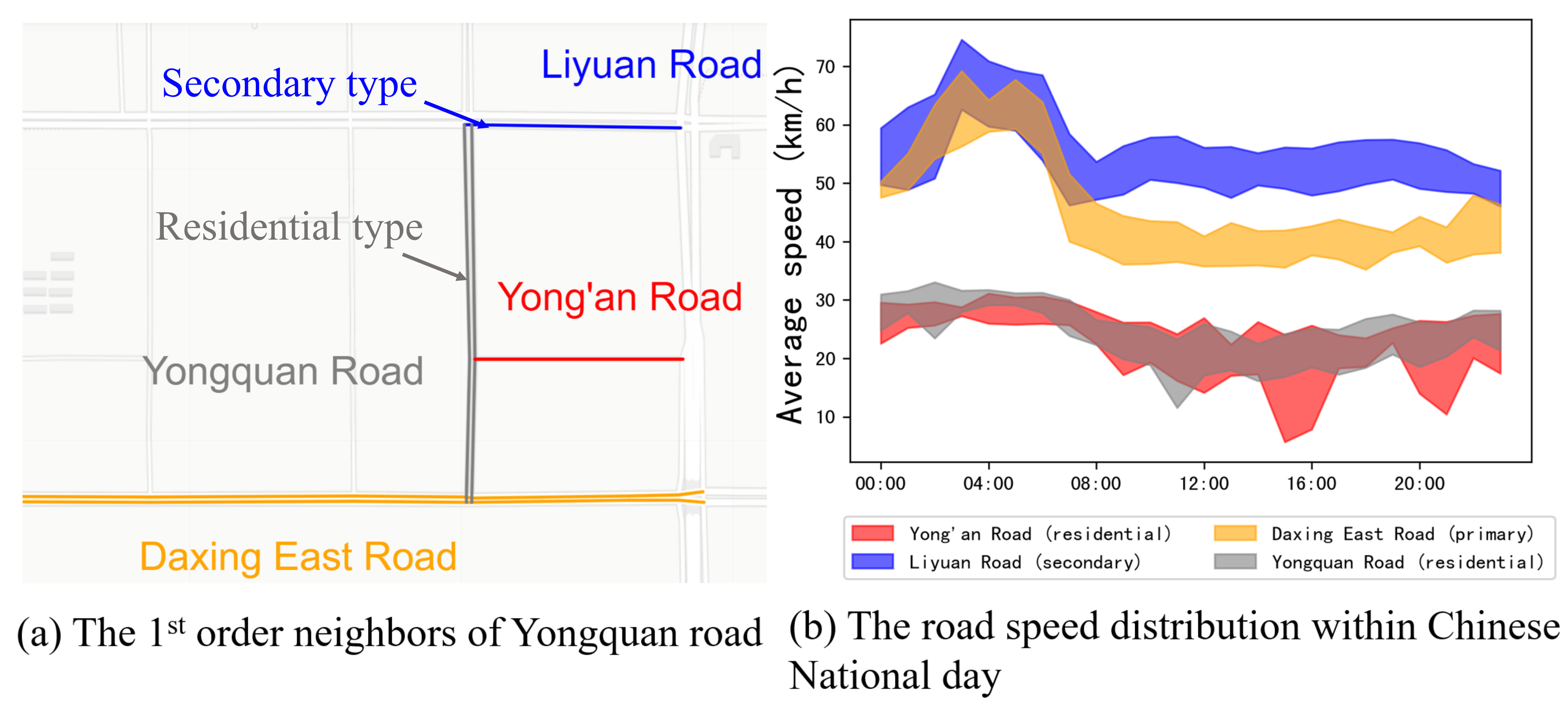}
	\caption{Motivation example of the traffic speed in Xi'an during Chinese National Day.  We used high-sampling-rate GPS trajectories to calculate the exact travel speed for these road segments, and observed that the road's speed distribution with  adjoining roads is highly related with the road types.}\label{fig:example}
\end{figure}

\subsection{Spatio-Temporal Model}

The spatio-temporal model aims to learn $f$ to estimate travel time distribution $Z$ of road segments from road network.  \figref{fig:example}(a) shows the locations of the neighbors of Yongquan Street, and  \figref{fig:example}(b) depicts the maximum, minimum and average traffic speed during Chinese national day. We observe several phenomena: 1) road speed is highly related to the road types; and 2) the road speed is also affected by the type of connection road. For example,  Yongquan road and  Liyuan street are residential and secondary roads, respectively. Even though they are neighbors, the road speeds are relatively different. Thus, the road network is represented as a multi-relational graph $\mathcal{G}=(\mathcal{V},\mathcal{E},\mathcal{R})$, where $\mathcal{V}$ denotes the set of vertices (i.e. road segments) and $\mathcal{E}$ denotes the set of edges. An edge $\mathbf{e}_{ijk}=(v_i,v_j,\mathbf{r}_k)\in \mathcal{E}$ indicates that the road segment $v_i\in \mathcal{V}$ connects to the road segment $v_j\in \mathcal{V}$ with a relation type $\mathbf{r}_k\in \mathcal{R}$.

Based on this multi-relational graph, we adopt a 3-layer Relational Graph Convolution Networks (R-GCNs)  \cite{schlichtkrull2018modeling} as the building block to learn the graph structure information.  $l^{th}$ R-GCN can be defined as 
\begin{align}
h_{v_i}^{(l+1)}=\sum_{\mathbf{r}_k \in \mathcal{R}} \sum_{j \in \mathcal{N}_{i}^{\mathbf{r}_k}} \frac{1}{c_{i, \mathbf{r}_k}} W_{\mathbf{r}_k}^{(l)} h_{j}^{(l)}+W_{0}^{(l)} h_{v_i}^{(l)},
\end{align}
where $ h_{v_i}^{(l)}\in \mathbb{R}^{d^{(l)}}$ is the hidden state of road segment $v_i$ in the $l^{th}$ layer of the model with dimension $d^{(l)}$ and $W_{\mathbf{r}_k}^{(l)}, W_{0}^{(l)} $ present the learnable parameters. $\mathcal{N}_{i}^{\mathbf{r}_k}$ denotes the set of neighbor indices of node $v_i$ in relation to $\mathbf{r}_k\in \mathcal{R}$. $c_{i, \mathbf{r}_k}$ is the normalization constant. Note that $h_{v_i}^{(0)}$ is the spatial feature $X_{v_i}$ of road segment $v_i$. We use the embedding layer to encode the following statistical features:
\begin{itemize}
	\item[$\bullet$] Road types: for example, primary, primary link, secondary, secondary link;
	
	\item[$\bullet$] Number of lanes: how many marked traffic lanes;
	\item[$\bullet$] Whether it is one way or not.
\end{itemize}
The final output of R-GCNs, represented as $s_{v_i} \in \mathbb{R}^{D}$, where $v_i\in \mathcal{V}$. But spatial representation $s_{v_i}$ only provides the static representation, which could not show the temporally dynamic correlations for each road segment. 

As we mentioned previously, the spatio-temporal model $f$ is to estimate the mean value $\mu$ and variance $\sigma$ of Lognormal distribution for each road segment $v_i$ at each time step $t_s$. Intuitively, we encode the day-of-week and time-of-day of each time step $t_s$ into $\mathbb{R}^7$ and $\mathbb{R}^\mathcal{I}$ using one hot encoding, and concatenate them with the embedding of weather conditions and HolidayID (holiday or not). Then we use one-layer MLP to transform the above temporal embedding vector into a vector $s_{t_s}\in \mathbb{R}^{D}$, which is equal to the spatial representation $s_{v_i}$. To obtain the time-variant road segment representations, we fuse the above spatial representation and reconstruct temporal embedding vector: for each road segment $s_{v_i}$ at time step $t_s$, the spatio-temporal representation is defined as $F_{v_i,t_s}=s_{v_i}+s_{t_s}$, which contains both spatial road structure and temporal information.

Based on the spatio-temporal representation $F\in \mathbb{R}^{(|\mathcal{V}|*N_{t_s})\times D}$, where $N_{t_s}$ denotes the total number of time steps, $\mu_{t_s}\in \mathbb{R}^{|\mathcal{V}|}$ and $\sigma_{t_s}\in \mathbb{R}^{|\mathcal{V}|}$ for each time step $t_s\in \mathbb{R}^{N_{t_s}}$ are parameterized by two-layer MLPs with shared fused representation $F$.

\subsection{Model Training with EM Procedure}

Next, we introduce the learning procedure of estimated travel time distribution $Z$ through weakly supervised learning. The aforementioned \textit{ expected log-likelihood} in \equref{eqn:weakly}  defines the aggregate expectation from $Z$ to $T$. Here, we assume the distribution $T_{a,b}$ also under the  Lognormal distribution approximated by summation of all $Z_{1:K}$ on route $r$ \cite{dufresne2008sums} as
\begin{align*}
	 \quad T &\sim \text{Lognormal}\left(\mu:=\sum_{i=1}^{K} \mu^{(i)}_{t_s},  \sigma^{2}:=\sum_{i=1}^{K} (\sigma^{(i)}_{t_s})^{2}\right) 
\end{align*}
Thus, the term of expectation in \eqnref{eqn:weakly} can be derived to be 
\begin{align}\label{equ:travel_loss}
	L_{\mu,\sigma}&=\log p\left(\left\{x^{(i)}_{t_s}, z^{(i)}_{t_s}\right\}_{i=1}^{K} ; W\right)\nonumber \\
	&= \sum_{i=1}^{K} \frac{\left(z^{(i)}_{t_s}-\mu^{(i)}_{t_s} \right)^{2}}{2 \sigma^{2}_{t_s}}
	-\frac{1}{2} \log \left(2 \pi \sigma^{2}_{t_s}\right)\nonumber\\
	&\approx-\frac{\left(Q(Z_{1: K})-Q(\mu) \right)^{2}}{2 \sigma^{2}}
	-\frac{1}{2} \log \left(2 \pi \sigma^{2}\right).
\end{align}
Here, we introduce how to maximize the conditional probability in \eqnref{eqn:search}.  Given two continuous samples $p_a$, $p_b$, and the last-step parameters $W^{(i)}$ produced from \eqnref{equ:travel_loss}, our objective is to find the optimal route $\widetilde{r}$ with the travel time 
closest to the observation $T_{a,b}$. As mentioned in \cite{he2020human}, it is natural to assume that the route $\widetilde{r}$ is very likely to be among the top $m$-shortest paths between $p_a$ and $p_b$. Therefore, we utilize Yen's algorithm \cite{yen1971finding} to generate the candidate set $\Omega_{a,b}=\{r_1,r_2,\cdots,r_m\}$  
and the optimal route $	\widetilde{r} $ can be selected by 

\begin{align}\label{equ:update_route}
	\widetilde{r} = \arg \min_{r} \mid T_{a,b} - \sum_{e_i \in r_j}\mu^{(i)} \mid, \quad \forall r_j \in \Omega_{a,b}.
\end{align}

We perform \eqnref{equ:update_route} for every pair of continuous samples to update their corresponding route $\widetilde{r}$. Furthermore, to prevent extensive overlap in $\Omega_{a,b}$, this paper leverages the weighted Jaccard ($wJCD$) value to calculate each pair of routes $r_i, r_j\in \Omega_{a,b}$ referring to \cite{he2020human}. The EM algorithm will complete when it reaches the estimated variables $\mu$ and $\sigma$ convergence.
\section{EXPERIMENTS}

\begin{table*}[!b]
 		\centering
 		\footnotesize
 		\caption{Performance comparison for travel time estimation under three datasets. Here, the units of both RMSE and MAE are minutes, and the unit of MAPE is percentages (\%).} 
 		\vskip 0.1in
 		\setlength{\tabcolsep}{0.8mm}{\begin{tabular}{l l  c c c |c c c | c c c}
 				\toprule \multirow{2}{*}{Data}   & \multirow{2}{*}{Models}   & \multicolumn{3}{c}{2 mins} & \multicolumn{3}{c}{4 mins}  & \multicolumn{3}{c}{8 mins} \\
				\cline{3-5}  \cline{6-8} \cline{9-11}   
 				&& {\small RMSE} & {\small MAE } & {\small MAPE} & {\small RMSE} & {\small MAE } & {\small MAPE} & {\small RMSE} & {\small  MAE} & {\small MAPE} \\
 				\midrule
 				\multirow{8}{*}{\rotatebox[origin=c]{90}{Xi'an}}&DeepTTE& 2.89&1.74&14.89&3.88&2.57&15.7 &3.87&2.79&18.25 \\
 				&DeepGTT & 4.31&3.51&32.16&5.27&4.11&39.85&7.33&6.14&45.31\\
 					
 				&MVSTM &4.13 &2.59&15.42&4.45&3.68&29.37&5.62&3.84&26.51 \\
 				&MURAT&8.86&6.87&84.06&9.45&7.76&94.16&11.36&9.23&113.69\\
 				&T-GCN & 3.24&  2.00  & 14.78&   4.23&  3.04& 15.61& 4.43&  3.19& 18.37  \\
				&DCRNN& 3.20 &  1.96& 14.6&    4.18&  2.98& 15.43& 4.37&  3.13& 18.16\\
				&ConSTGAT &3.21&  1.99& 14.41&   4.20 &  3.01& 15.22& 4.39&  3.17& 17.91 \\
				&Ours&\textbf{1.36}&\textbf{1.07}&\textbf{9.66}&\textbf{1.53}&\textbf{1.32}&\textbf{13.35}&\textbf{1.81}&\textbf{1.92}&\textbf{15.69}\\ 
 				\midrule
 				\multirow{8}{*}{\rotatebox[origin=c]{90}{Porto}}
				&DeepTTE	&2.14&1.46&12.59&2.90&2.10&12.36&3.38&2.48&15.01  \\
				&DeepGTT 	&3.39&2.83&28.58&4.07&3.03&32.05&6.23&5.08&39.34  \\
				&MVSTM 		&3.10&2.27&13.00&3.39&2.75&24.28&4.69&3.18&22.18  \\
				&MURAT		&6.24&5.35&74.27&7.17&5.63&75.54&9.40&7.39&96.79  \\
				&T-GCN 		&2.04&1.31&12.13&2.85&1.83&11.92&3.16&2.23&14.26  \\
				&DCRNN		&2.02&1.28&11.99&2.81&1.80&11.78&3.12&2.19&14.09  \\
				&ConSTGAT   &2.03&1.29&11.83&2.82&1.82&11.62&3.13&2.22&13.90  \\
				&Ours&\textbf{1.11}&\textbf{0.88}&\textbf{8.25}&\textbf{1.30}&\textbf{1.14}&\textbf{10.58}&\textbf{1.72}&\textbf{1.63}&\textbf{12.77}\\ 
					\midrule

				\multirow{8}{*}{\rotatebox[origin=c]{90}{Chengdu}}&DeepTTE& 3.13&1.97&15.11&4.09&2.99&15.96 &4.28&3.14&18.79 \\
				&DeepGTT & 4.97&3.83&34.31&5.75&4.32&41.39&7.89&6.44&49.24\\
				
				&MVSTM &4.54 &3.07&15.61&4.78&3.92&31.36&5.94&4.03&27.77\ \\
				&MURAT&9.13&7.24&89.17&10.12&8.03&97.57&11.92&9.36&121.16\\
				&T-GCN & 2.99&  1.77  & 14.56&  4.02&  2.61& 15.39& 4.01& 2.83& 17.85  \\
				&DCRNN& 2.95&  1.73& 14.39&    3.97&  2.56& 15.21& 3.96&  2.78& 17.64\\
				&ConSTGAT &2.97&  1.75& 14.20&   3.98 &  2.59& 15.01& 3.97&  2.81& 17.40 \\
				&Ours&\textbf{1.89}&\textbf{1.31}&\textbf{10.15}&\textbf{2.14}&\textbf{1.93}&\textbf{13.98}&\textbf{2.55}&\textbf{2.21}&\textbf{16.28}\\ 
 				\bottomrule
 		\end{tabular}}

 		\label{table:result1}
 	\end{table*}	
\subsection{Experimental Settings}

\subsubsection{Data.} We validate our proposed methods on three real-world datasets,
including Xi'an, Porto and Chengdu dataset. More details can be found in
Appendix A.1. For each dataset, we use the trajectories generated in the last 7 days as the test set and the rest of the trajectories as the training set. To avoid the influence of a very small number of trajectories during late-night hours, we conducted the travel time estimation and route recovery for taxi trajectory experiments from 6:00 AM to 22:00 PM. 

\subsubsection{Sampling Rate Setting.} According to \cite{yuan2010interactive}, taxis should report their GPS positions with a low sampling rate to save communication and energy costs. We further vary the sampling ratio of the sets 3.125\%, 6.25\% and 12.5\% to evaluate the robustness of our proposed model. Since the original trajectories are sampled every 15 seconds, the generated low- sampling-rate trajectories of 3.125\%, 6.25\% and 12.5\% are considered to be as the average time interval of such trajectories is 8 mins, 4 mins, and 2 mins, respectively.

\subsection{Baseline Models \& Evaluation Metrics} 

\begin{itemize}
	\item[$\bullet$] \textbf{DeepTTE:} \cite{wang2018will} is an end-to-end deep learning framework, which infers the  travel time  from both the entire path and each local path simultaneously.
	\item[$\bullet$] \textbf{DeepGTT:} \cite{li2019learning} learns the travel time distribution through the deep generative model, which takes the real-time traffic condition into account.
	 
	 \item[$\bullet$] \textbf{MVSTM:} \cite{liu2021multi} is a multi-view spatial-temporal model that captures the mutual dependence of spatial-temporal relations and trajectory features. 
	 
	\item[$\bullet$] \textbf{MURAT:} \cite{li2018multi} is a multi-task representation learning method by utilizing the underlying
	road network and the spatio-temporal prior knowledge.
	
	\item[$\bullet$] 
	\textbf{DCRNN:}  \cite{li2017diffusion} exploits GCN to capture spatial dependency and then uses recurrent neural networks to model temporal dependency.

\item[$\bullet$] 
\textbf{ConSTGAT:}  \cite{fang2020constgat} adopts a graph attention mechanism to explore the joint relations of spatio-temporal information.
	
\item[$\bullet$] 
\textbf{T-GCN:}  \cite{zhao2019t}  proposes a temporal GCN model that combines the GCN and GRU to simultaneously extract the spatial and temporal dependencies.
	
\end{itemize}

Additionally, three state-of-the-art algorithms STRS \cite{wu2016probabilistic}, MTrajRec \cite{ren2021mtrajrec}, and DeepGTT \cite{li2019learning} are used as baseline models for route recovery. STRS and DeepGTT  learn the travel time of the road network and conduct the route search for  low-sampling-rate trajectories.
MtrajRec recoveries the route via a two-stage Seq2Seq model based on coarse grid representation.

\noindent\textbf{Evaluation Metrics.} We evaluate the task of TTE with RMSE (radial mean square error), MAE (mean absolute error) and MAPE (mean absolute percentage error). Then the route recovery performance is evaluated by route recovery accuracy, which is defined as the ratio of the length of correctly inferred road segments against the maximum value of the length of the ground truth route $R_G$ and the inferred route $R_I$, that is, $accuracy = \frac{(R_{G}\cap R_{I}).len} {max\{R_{G}.len,R_{I}.len\}} $.

\subsection{Performance Comparison}

\noindent\textbf{Performance on Travel Time Estimation.} As reported in \tableref{table:result1}, our WSL-TTE achieves the best results among all baseline methods for three kinds of minute intervals: {2, 4, 8}. We summarize the reasons for our model outperforming all baselines by a large margin: 1) The GCN production is used to learn the route's travel time and its travel time distribution for each road segment simultaneously, helping it to yield robust and abundant features. 2) The EM iteration algorithm has been proposed to update the potential travelled path, which helps our method to learn a more reasonable travel time distribution. Meanwhile, the estimation results among different sampling intervals also reflect that the uncertainty of a sparse GPS trajectory would seriously affect the model performance.  
\begin{table}[t]
 		\centering
 		\small
 		\caption{Performance of WSL-TTE and its variants for travel time estimation under extremely sparse scenario (8 mins).} 
 		\vskip 0.1in
 		\setlength{\tabcolsep}{1.2mm}{\begin{tabular}{l  c c c }
 				\toprule \multirow{2}{*}{Models}   & \multicolumn{3}{c}{Xi'an/Porto/Chengdu}  \\
				\cline{2-4}  
 				& {\small RMSE} & {\small MAE } & {\small MAPE} \\
 				\midrule
 				simpleGCN		& 1.92/2.09/2.83&	2.04/1.74/2.37&15.96/13.09/16.82 \\
 				GAT 			& 1.96/2.14/2.89&	2.15/1.80/2.42&15.92/13.10/16.87\\
 				 GTN			&1.95 /2.10/2.82 &	2.11/1.78/2.41&15.94/13.07/16.77 \\ \hline
 			Normal Distribution & 1.88/2.06/2.79&	2.09/1.84/2.56&15.88/13.09/16.89\\
 				 Variance=1		&2.07 /2.43/3.45 &	2.71/2.17/2.79&16.12/13.28/17.13 \\
 				
 			\hline	Ours		&\textbf{1.81/1.72/2.55}&\textbf{1.92/1.63/2.21}&\textbf{15.69/12.77/16.28} \\ 
                
 				\bottomrule
 		\end{tabular}}

 		\label{table:ablation}
 	\end{table}

\begin{figure}[!b]
    \centering
    \subfigure[Route recovery accuracy of Xi'an.]{
        \label{figure:5a}
        \includegraphics[height=0.25\linewidth]{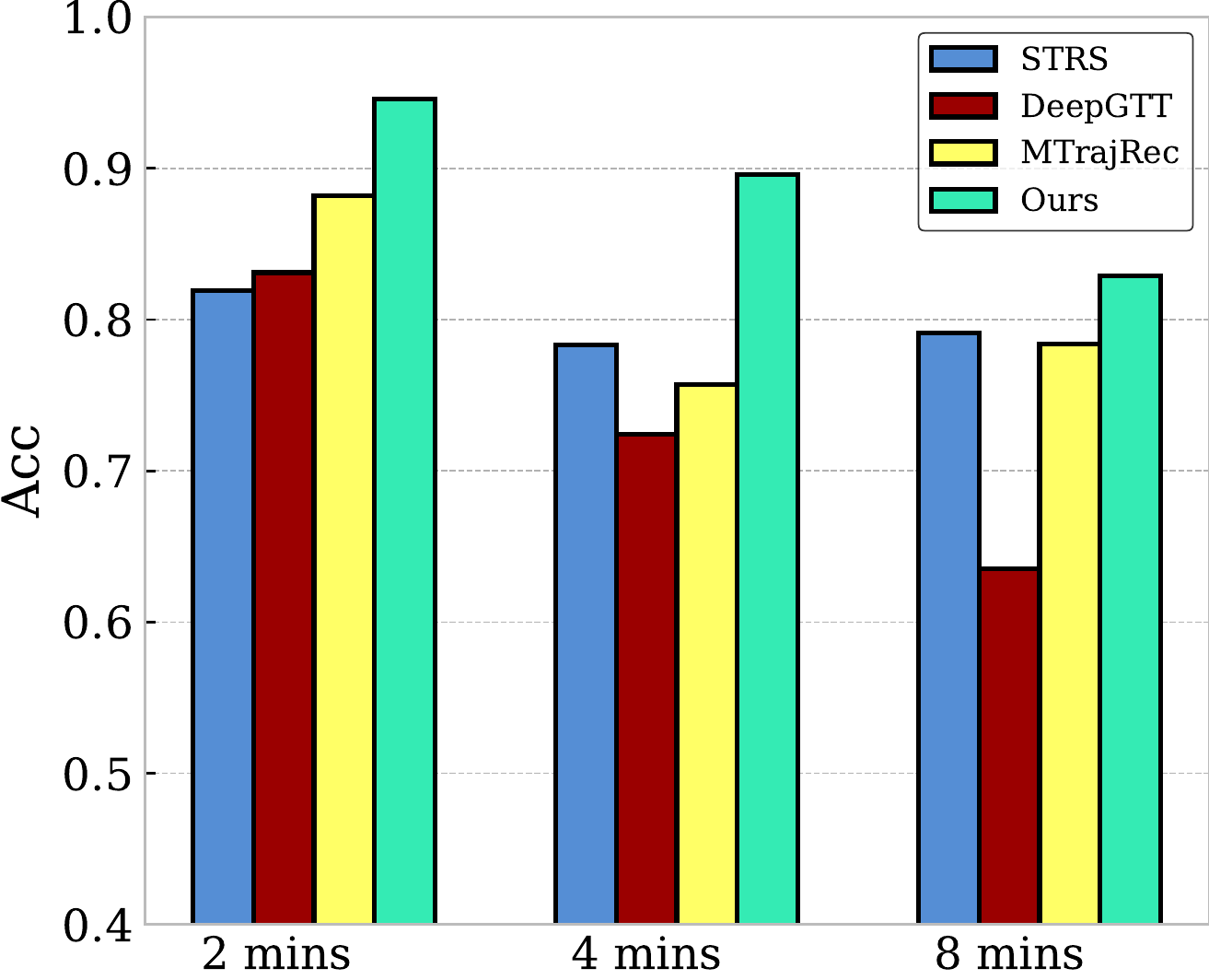}
    }
    \hfill
      \subfigure[Route recovery accuracy of Porto.]{
        \label{figure:5b}
        \includegraphics[height=0.25\linewidth]{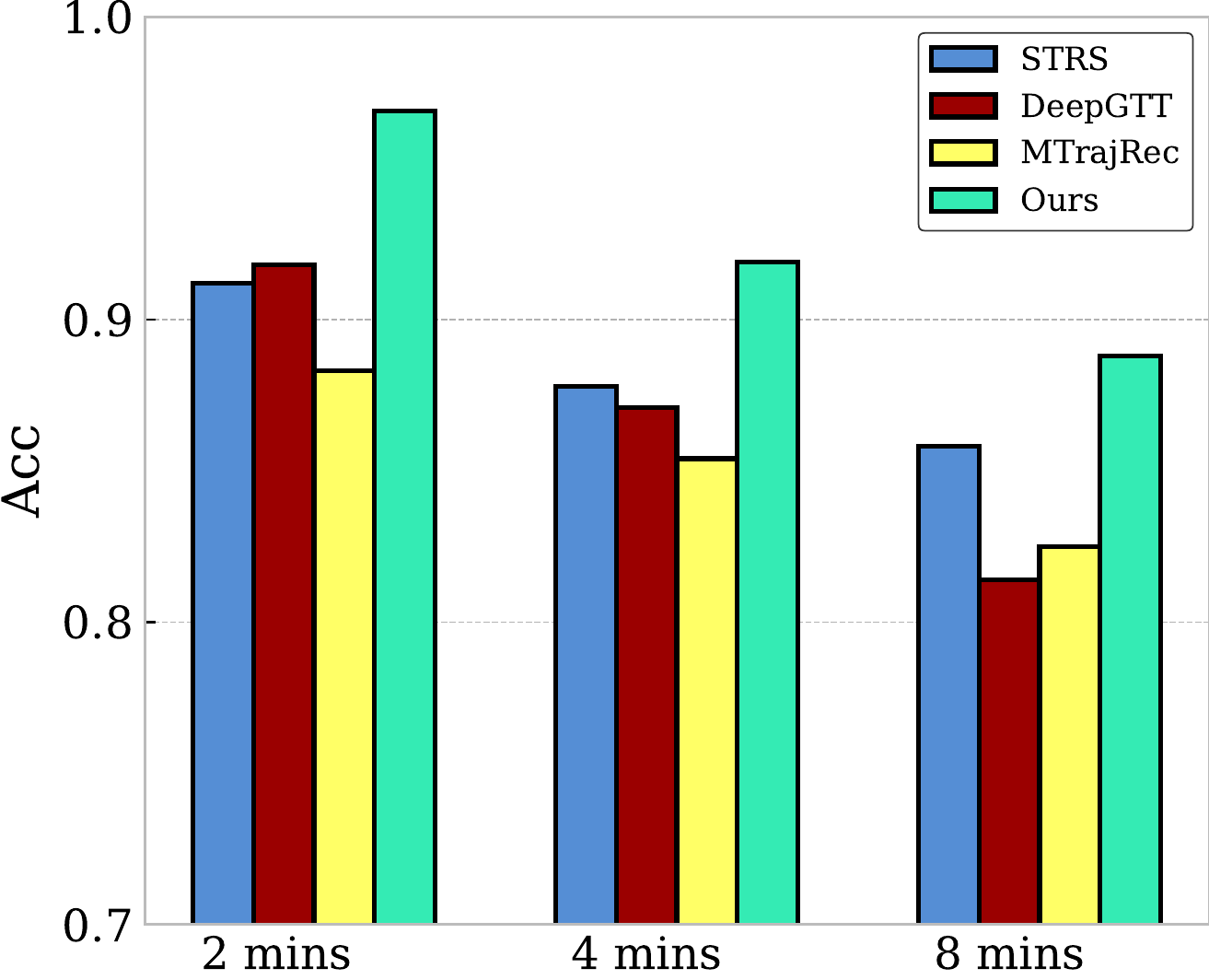}
    }
    \hfill
      \subfigure[Route recovery accuracy of Chengdu.]{
        \label{figure:5c}
        \includegraphics[height=0.25\linewidth]{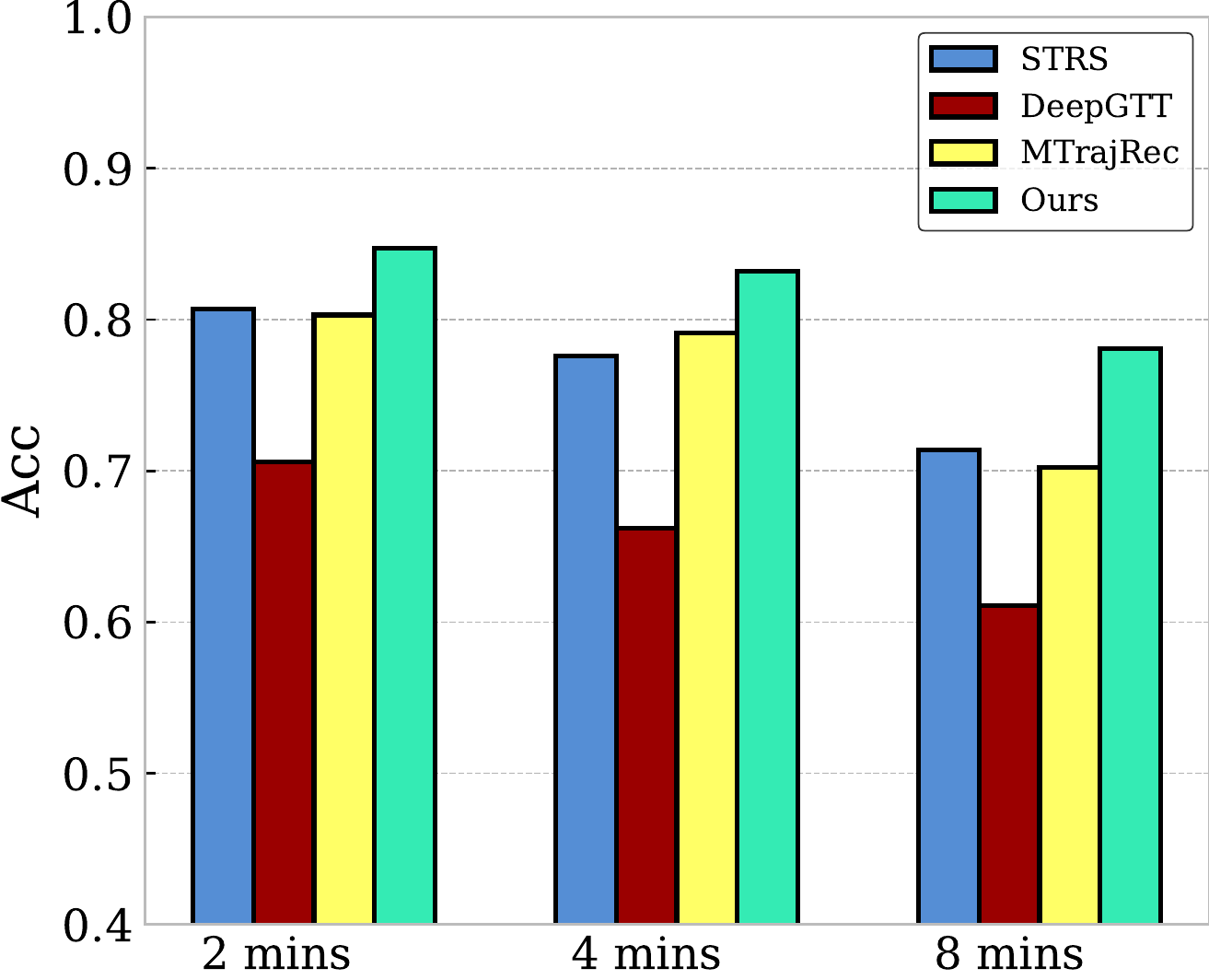}
    }

      \subfigure[Daily divergence of Xi'an by ours.]{
        \label{figure:5d}
        \includegraphics[height=0.3
        \linewidth]{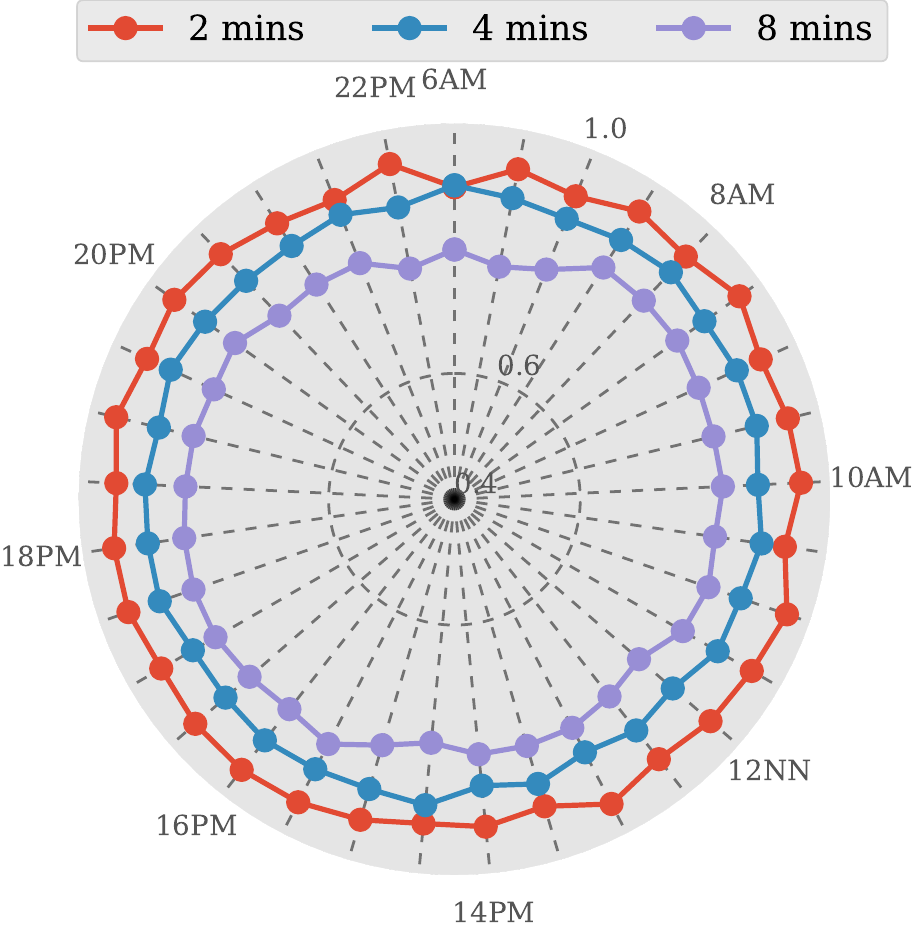}
    }
    \hfill
        \subfigure[Daily divergence of Porto by ours.]{
        \label{figure:5e}
        \includegraphics[height=0.3
        \linewidth]{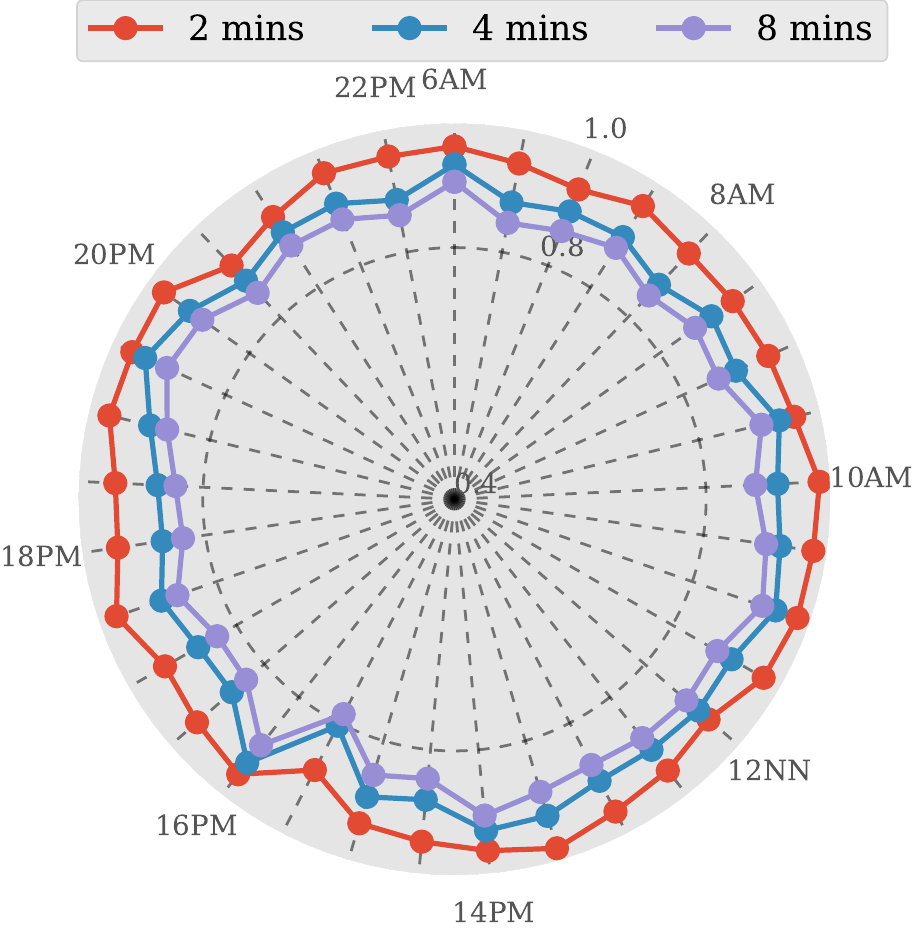}
    }
    \hfill
    \subfigure[Daily divergence of Chengdu by ours]{
        \label{figure:5f}
        \includegraphics[height=0.3
        \linewidth]{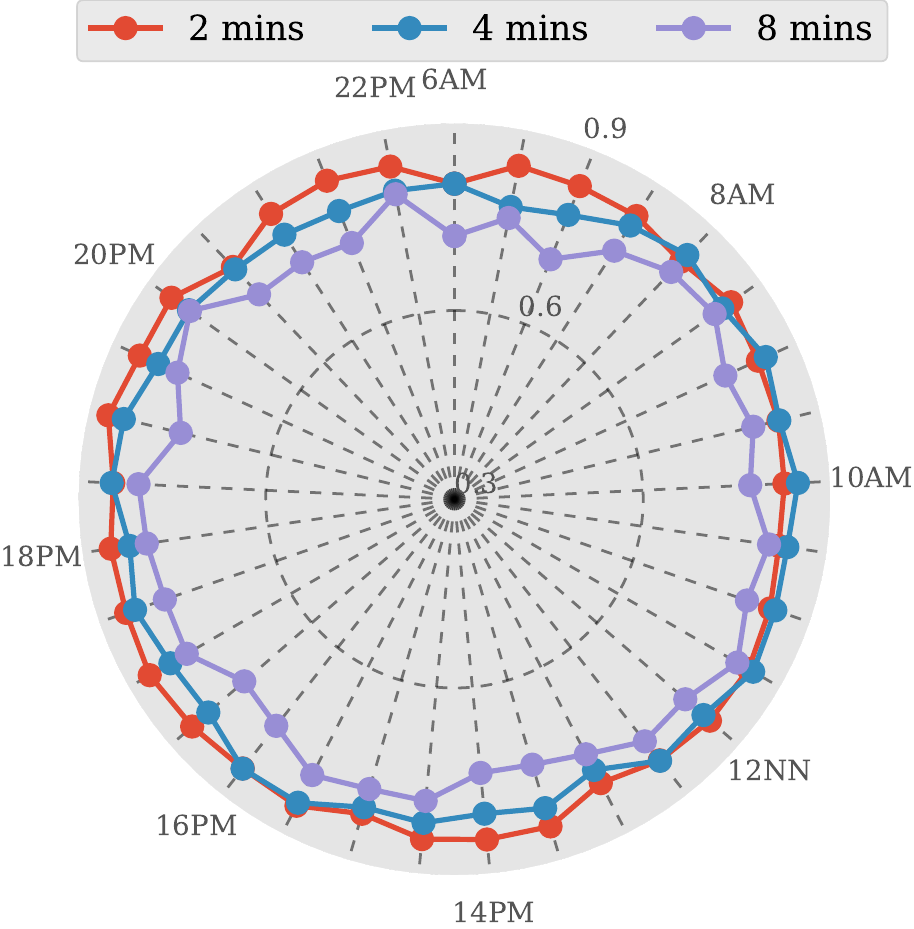}
    }
    \caption{Route Recovery Performance.}
    \label{figure:route}
\end{figure}
\noindent\textbf{Ablation Study.} As is shown in \tableref{table:ablation}, in order to validate how the relational GCN modules and weak supervision can effectively capture the spatio-temporal dependencies in WSL-TTE,  we first test the effects of relational GCN on modeling road network. Our WSL-TTE removes the relational GCN module and replaces it with a simple GCN, GAT, \cite{velivckovic2017graph} and graph transformer network \cite{yun2019graph}  to extract the spatial representation. The experimental results show that using our model with relational GCN can achieve better performance on two datasets. It can be explained that the complex adjacency of the road network needs to model different correlations among different road types. Next, we validate the assumption that the travel time variables belong to the Lognormal distribution. Compared to this setting, we conduct the test of normal distribution and variance=1, respectively. We find that they cannot achieve better performance than the Lognormal distribution. This validates our formulation of weak supervision regarding travel time. In sum, we can conclude that our 
WSL-based method is effective in travel time estimation.

\noindent\textbf{Performance on Route Recovery.} 
\figref{figure:route} reports the route recovery accuracy and daily divergence over different sampling time intervals. \figref{figure:5a}, \figref{figure:5b} and \figref{figure:5c} show that our WSL achieves better performance among three sampling intervals, compared to DeepGTT, MTrajRec and STRS. Noted that DeepGTT achieves worse performance at 4 mins and 8 mins. This is because grid-based traffic condition tensors can not provide efficient road conditions at high sampling intervals. \figref{figure:5d}, \figref{figure:5e} and \figref{figure:5f} provide the daily divergence of the route recovery accuracy by our proposed WSL. We find that the total recovery performance stays stable from 6:00 AM to 22:00 PM for Xi'an, Porto, and Chengdu. However, the performance of Chengdu is relatively worse than both Xi'an and Porto due to the more complex road network. Furthermore, significantly as the sampling time interval increases, the accuracy of both methods drops, as expected. The reason is that a more extensive sampling time interval leads to more possible candidate routes to be inferred between two sample points. Meanwhile, the daily divergence of our WSL also shows this pattern.


\begin{figure}[!b]
	\centering
		\centering
	\includegraphics[width=0.8\linewidth]{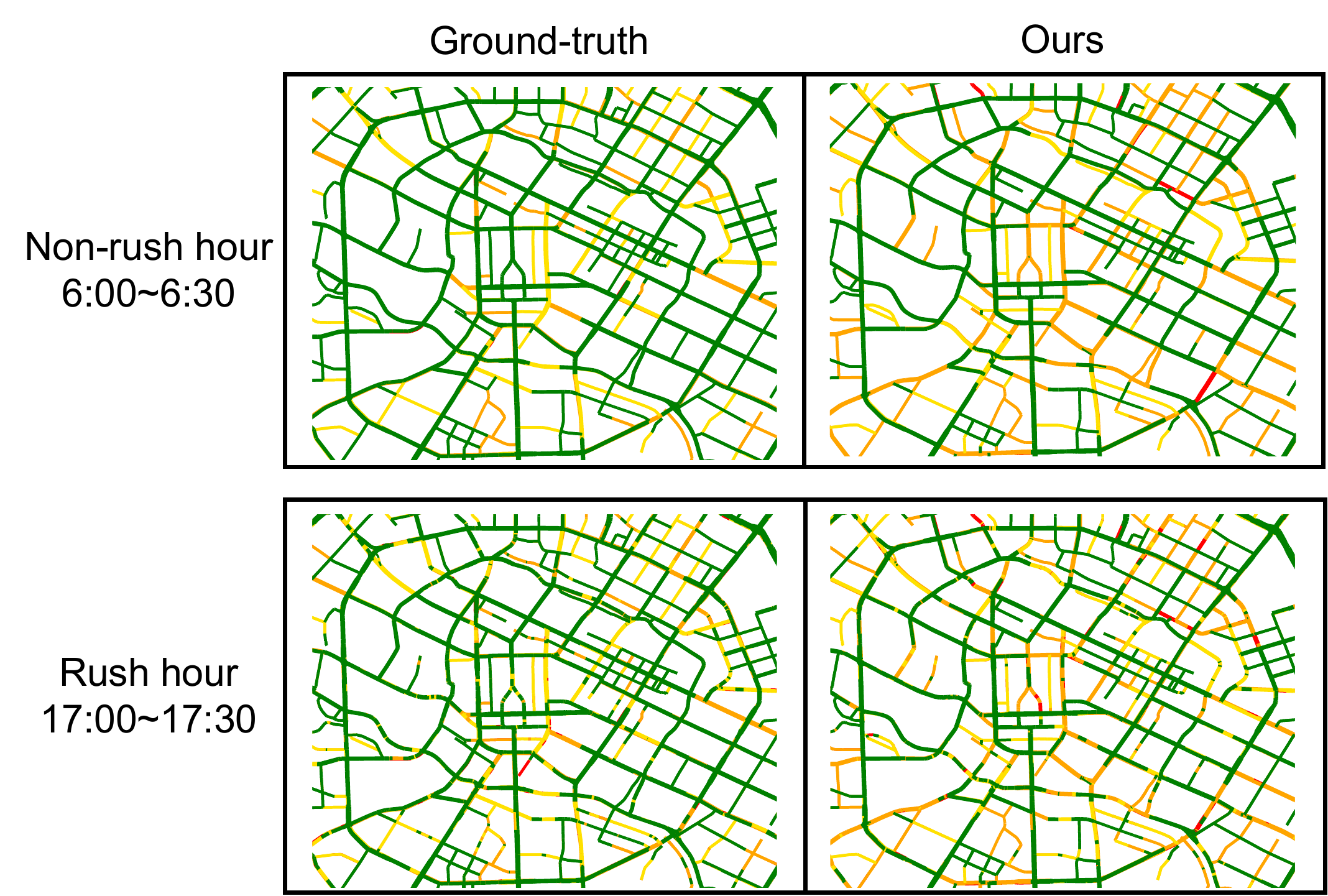}
	
	\caption{Traffic condition comparison. We pick two-time steps, i.e., non-rush hour (6:00-6:30)
and rush hour (17:00-17:30), and compute the ground truth by original dense trajectories  in Chengdu, compared with the transformed speed based on the learned travel time distributions of our WSL-TTE. }
	\label{figure:road_con}
\end{figure}

\subsection{Case Study}

We conducted a real-world case study in Chengdu, which visualizes the learned road conditions using our proposed WSL-TTE. To acquire the road conditions of the road network, we here transform travel time $\mu$ estimated by our WSL-TTE into the average speed by $Speed_i=\frac{Length_i}{\mu_i}$ for each road $v_i$. Four kinds of colors are used to represent the different road states, which can be defined as 1) red - very congested, 2) yellow - congested, 3) orange - slow, and 4) green - unblocked. We equally divide the limiting velocity for each road type and set the speed interval for these four road states. For example, the speed limit of the primary road type is $60kph$, then the speed range that represents very congested is $[0, 15)$, congested is $[15, 30)$, slow is $[30, 45)$ and unblocked is $[45, 60)$. We calculate the average speeds by the original dense trajectories as the ground truth. Specifically, we mark them with an unblocked state for the roads without a trajectory. The compared result is shown in \figref{figure:road_con}, our model can generate approximate road conditions with ground truth for both non-rush and rush hour.

\begin{figure}[h]
	\centering
	\includegraphics[width=0.8\linewidth]{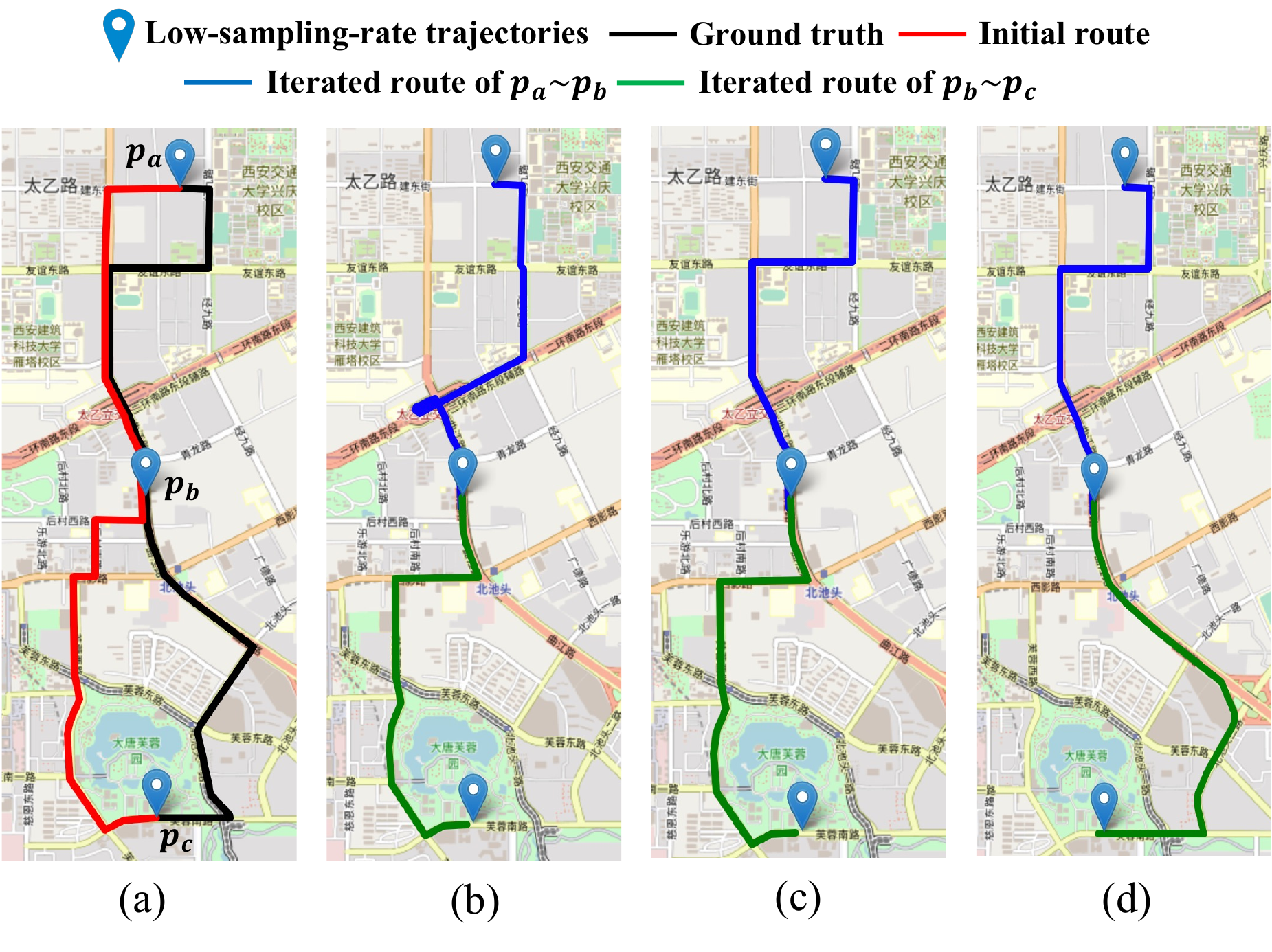}
	\caption{ Visualization example of the route update process. Here, Fig. (a) shows the ground truth and initial route of low-sampling-rate trajectories, and Fig. (b)$\sim$(d) shows the intermediate iteration results of route recovery. }\label{fig:visual_example}
\end{figure}

 In addition, we provide a visualization example of the route update process in route recovery. As is shown in \figref{fig:visual_example}, our route update process with the EM algorithm can gradually find the approximate route with ground truth, owing to the precise travel time, which demonstrated our previous assumption that: the more precise travel time can lead to a better inference of routes, in turn, resulting in more accurate time estimation. 

\section{CONCLUSION}


This paper formulates the TTE and route recovery in a highly sparse scenario as
an inexact supervision. Based on the EM algorithm,  we solve the inexact travel time labeling and uncertain route choice  by alternatively updating the travel time distribution through weakly supervised learning and route searching using the top $m$-shortest path respectively. Both two tasks are complementary to each other in the iteration process. In future work, we intend to consider more hypotheses of travel time distribution under weakly supervised learning, such as Gamma, Weibull, as well as Burr XII distribution.  Source code is available at https://github.com/Dreamzz5/WSL\_TTE. 

\section*{Acknowledgment}
This work was partially supported by National Key Research and Development Project (2021YFB1714400) of China and  Guangdong Provincial Key Laboratory (2020B121201001).

\bibliographystyle{splncs04}
\bibliography{paper}

\section*{A APPENDIX}

\setcounter{figure}{0}    
\setcounter{table}{0}
\setcounter{equation}{0}

\subsection*{A.1 Datasets} 
\noindent\textbf{Road Networks.} The matching road network database has been collected from OpenStreetMap for three regions, including Porto, Xi'an and Chengdu. Here, the Porto road network possesses 10334 edges and 9072 nodes, the Xi'an road network ranges from 34.20\degree to 34.29\degree in latitude and 108.90\degree to 108.99\degree in longitude, and it
contains 4780 edges and 3782 nodes, and the Chengdu road network 
contains 8221 edges and 5182 nodes. All three  road networks included nine road types (trunk, trunk link, freeway link, primary, primary link, secondary, secondary link, tertiary, tertiary link). In this paper, we use the Fast Map Matching (FMM) algorithm \cite{yang2018fast} to map the vehicle trajectory onto the road network.

\noindent\textbf{Taxi Trajectories.} We use three real-world taxi datasets to evaluate our proposed method. Porto\footnote{https://www.kaggle.com/c/pkdd-15-predict-taxi-service-trajectory-i/data} dataset contains the taxi trajectories of 442 taxis in May 2014. Xi'an and Chengdu come from Didi Chuxing Open Data, which are October 2016 and August 2014 respectively. The GPS points of both two datasets have been tied to the road and the interval of sample trajectory points is 2-4s. 
Especially, we further collected the corresponding weather conditions \footnote{https://www.timeanddate.com/weather} ( 6 types including temperature, wind, visibility, etc.) of each trajectory for all three datasets.

\subsection*{A.2 Hyper-parameter Setting} 

The parameter settings used in our experiment are described as follows:

\begin{itemize}
	\item[$\bullet$] In the generation of a candidate route set $\Omega_{a,b}$, $m$ candidate routes are selected between the two given low sample rate GPS points. Here, $m=5$ is used for both Xi'an and Porto, and $m=7$ for Chengdu, respectively. All three hyper-parameters can ensure that over 90 percents of ground truth route can be acquired from $\Omega_{a,b}$.
	\item[$\bullet$] In the spatial R-GCN layers, the embedding sizes of road feature representation (road types, number of lanes and one way or not) are set to 8, 4, 2 respectively. 
	\item[$\bullet$] In the temporal embedding component, we embed WeatherID and HolidayID in $R^8$ and $R^4$, respectively.
\end{itemize}

 The experiments are implemented with PyTorch 1.6.0 and Python 3.6 and are trained with a RTX2080 GPU. The platform ran on Ubuntu 16.04 OS. Our model is optimized by Adams with an initial learning rate 0.0001 for 20 epochs, and early stopping is used on the test dataset. We select the best hyper-parameter setting for the baseline methods on the valuation dataset. 
 
 \subsection*{A.3 Case study for Learned Travel Time Distribution}

\begin{figure}[h]
    \centering
    \subfigure[Xiaozhai east road (type: ``primary'', length: 458m).]{
        \label{figure:61a}
        \includegraphics[height=0.25\textwidth]{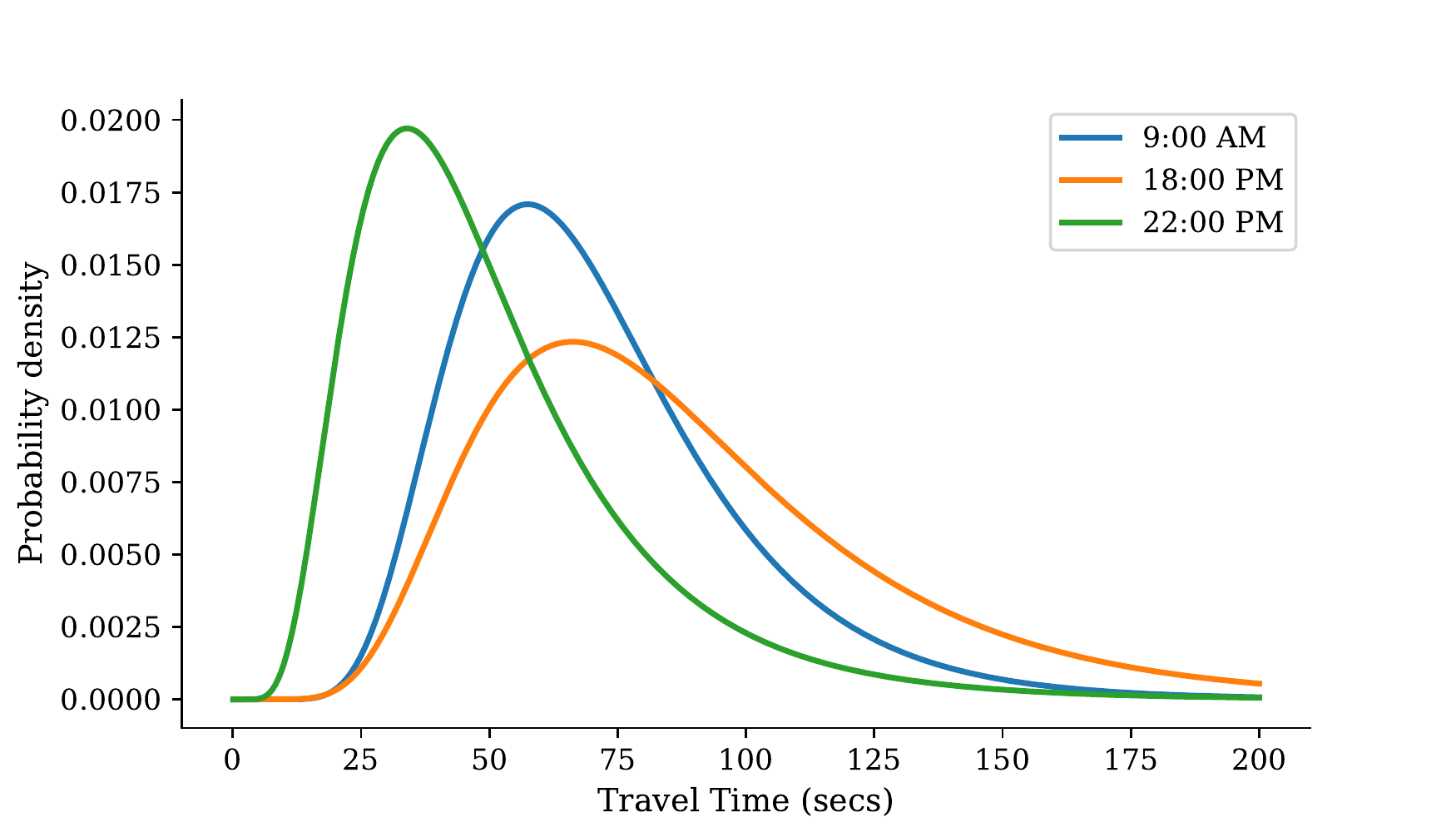}
    }
    \subfigure[East part of second ring south road (type: ``trunk'', length: 662m).]{
        \label{figure:62b}
        \includegraphics[height=0.25
        \textwidth]{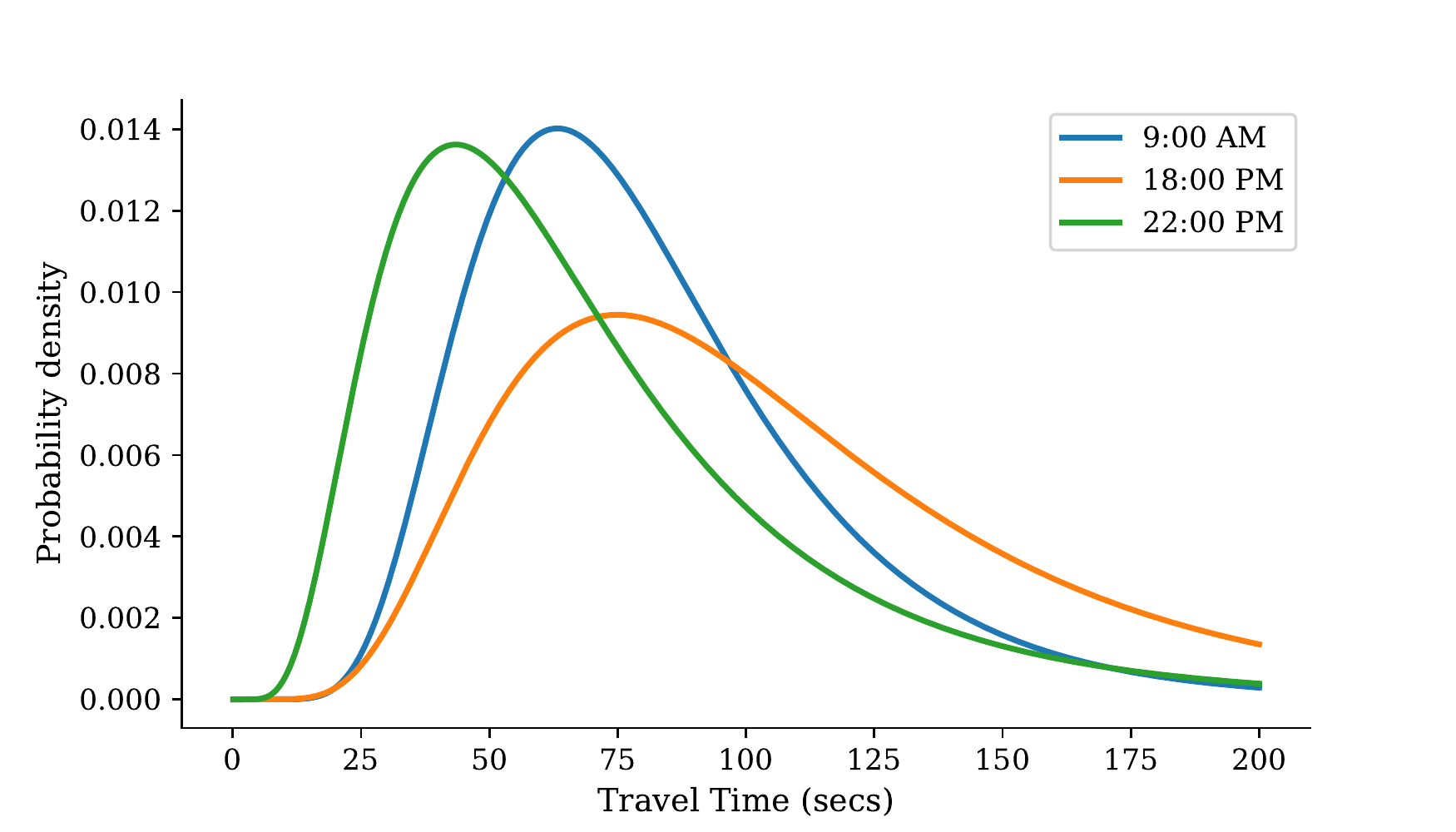}
    }

    \caption{Learned travel time distributions by our WSL-TTE. }
    \label{figure:road_dis}
\end{figure}

In addition, to validate the effectiveness of the learned travel time distributions of road segments, we also select two road segments from the Xi'an road network with the similar travel time. One is a busy road segment (type: ``primary''), the other is a highway. From \figref{figure:road_dis}, we can find that learned travel time distributions for both two roads have the obvious morning and evening rush hours. Meanwhile, this primary road has a relatively more considerable variance than the highway. This demonstrates that our WSL can acquire reasonable travel time distributions of road network.

\end{document}